\documentclass[11pt]{article}
\usepackage[final]{acl}
\usepackage{times}
\usepackage{latexsym}
\usepackage[T1]{fontenc}
\usepackage[utf8]{inputenc}
\usepackage{microtype}
\usepackage{inconsolata}
\usepackage{graphicx}
\usepackage{booktabs}
\usepackage{tabularx}
\usepackage{array}
\usepackage{amsmath}
\usepackage{xspace}

\newcommand{\method}[1]{\texttt{#1}}
\newcolumntype{Y}{>{\raggedright\arraybackslash}X}
\hypersetup{
  pdftitle={When Learned Context Planning Fails to Beat Strong Retrieval: A Controlled Study of Planning, Routing, and Reranking for Long-Context QA},
  pdfauthor={Yingrui Li and Han Chen}
}

\title{When Learned Context Planning Fails to Beat Strong Retrieval:\\
A Controlled Study of Planning, Routing, and Reranking for Long-Context QA}

\author{
\textbf{Yingrui Li}\textsuperscript{1} \quad
\textbf{Han Chen}\textsuperscript{2} \\
\textsuperscript{1}Independent Researcher, United States \quad
\textsuperscript{2}Independent Researcher, United States \\
\small{\texttt{emmalyr0404@gmail.com}, \texttt{hc414t@gmail.com}}
}

\begin{document}
\maketitle
\begin{abstract}
Learned context planning selects evidence atoms before an answer model reasons over them. We test whether this learned selection improves long-context multiple-choice QA after strong retrieval, routing, budgeted-selector, and reranking controls. Our primary diagnostic uses all 503 LongBench-v2 MCQ questions with Qwen2.5-7B-Instruct. The planner is SFT-trained on outcome-selected traces from 140 training and 28 development questions; because the 503-question analysis includes those questions, it is partly transductive. At an 18k-character budget, anchored hybrid retrieval reaches 36.18\% accuracy and BM25 reaches 35.98\%, while the best direct planner-guided method reaches 34.19\%. On the untouched 152-question test split, anchored hybrid remains higher (42.11\% versus 36.84\%). Leakage-safe routers cannot convert a large oracle gap. Under tight budgets, the best planner is ahead by only 0.40 points at 6k and loses at 9k; planner-guided reranking has a +1.79-point estimate at 6k with a paired interval crossing zero and ties the control at 9k. Packing-order and score-flatness analyses did not identify a stable mechanism. Under this setup, learned planning is a weak relevance signal rather than a replacement for strong retrieval.
\end{abstract}

\section{Introduction}
Long-context question answering systems must decide which parts of a large input to show to an answer model. Retrieval-augmented generation (RAG) usually ranks chunks with lexical, dense, or hybrid retrieval, then packs the top results into context \citep{robertson2009probabilistic,lewis2020retrieval,reimers2019sentence}. A learned context planner offers a different interface: retrieve a candidate set, ask a trained model to select evidence atoms, and answer from the selected or planner-prioritized context. Here, planning means evidence selection; the planner outputs atom IDs rather than multi-step reasoning traces.

This idea is plausible, but a weak comparison can make it look stronger than it is. A planner may beat prompted planning or a direct baseline while still losing to BM25, and a large oracle routing bound may not imply a deployable inference-time router.

We ask: \emph{when does learned evidence planning survive strong retrieval and reranking controls?} We answer this question on LongBench-v2 multiple-choice QA \citep{bai2024longbenchv2}. Learned planning changes the evidence set, helps some examples, and produces a small 6k-character rerank signal. However, in this setting it does not robustly beat strong retrieval, routing, budgeted selection, or reranking controls.

We show that (i) direct planner-guided construction loses to BM25 and anchored hybrid retrieval, (ii) oracle complementarity does not become a leakage-safe router, (iii) low-budget planner gains shrink under stronger selectors, and (iv) the remaining planner-rerank signal is statistically unsupported. We additionally audit train--evaluation separation, held-out behavior, domain variation, packing order, and retrieval-score mechanisms. The practical takeaway is that planner gains should be checked against strong retrieval, routing, budgeted selection, and reranking controls before being attributed to planning.

\section{Experimental Setup}
\paragraph{Task, split, and executor.}
We deterministically split the 503 LongBench-v2 MCQ questions into 301 training, 50 development, and 152 test questions (split seed 20260225). Each example has a long context, question, choices, and a gold letter; a run is correct when the parsed output letter matches the gold. The answer model is Qwen2.5-7B-Instruct \citep{qwen2024qwen25}. Our primary diagnostic reports all 503 questions and therefore includes the 168 questions used to construct planner SFT train/development records. It is partly transductive and is not a conventional held-out planner-generalization estimate. We separately report the untouched 152-question test partition.

\paragraph{Atoms and retrieval controls.}
An \emph{evidence atom} is a deterministic character-bounded chunk. Multi-document contexts are split on document separators or headers, then chunked into 2,000-character atoms with 200-character overlap; other contexts use 8,000-character atoms with 1,000-character overlap. A method ranks or selects atoms and packs their full text under an 18k, 9k, or 6k character budget. BM25 uses an alphanumeric tokenizer. Dense retrieval uses \method{sentence-transformers/all-MiniLM-L6-v2}; hybrid retrieval applies reciprocal-rank fusion. \emph{Anchored hybrid} places top BM25 atoms before hybrid fill. Budgeted controls include no-anchor and weighted hybrid packing and MMR. Non-planner rerankers score the same BM25+dense candidate pool with \method{cross-encoder/ms-marco-MiniLM-L-6-v2} \citep{wang2020minilm}.

\paragraph{Trace mining and planner SFT.}
For each training or development question, the base Qwen model samples six plans from a 30-atom retrieval candidate pool (temperature 0.7, top-$p$ 0.9, at most eight selected atoms, and an 18k-character execution budget). We deduplicate plans by selected atom set, execute each unique plan with the same base answer model, and assign binary reward from answer correctness. For SFT, we retain one reward-one plan per question, preferring fewer atoms, shorter context, and earlier sample order. This yields 140 SFT training rows and 28 development rows, with zero test-ID overlap.\footnote{The submission version reported 262 retained traces in a merged SFT archive. A release audit could not reproduce that count from the audited artifacts of the final adapter; the verified counts are the 140 training and 28 development rows (168 unique questions) reported here.} We train a Qwen2.5-7B-Instruct LoRA adapter \citep{hu2022lora} for three epochs with rank 16, alpha 32, dropout 0.05, learning rate $5\times10^{-5}$, per-device batch size 2, gradient accumulation 8, bfloat16 precision, and seed 111, using Transformers and PEFT \citep{wolf2020transformers,mangrulkar2022peft}. The final adapter is the global-step-24 state; no development checkpoint selection occurs because training ends before the configured evaluation interval.

At inference, the planner receives only the question, choices, and retrieved candidate atoms. \emph{Plan-guided hybrid} packs planner-selected atoms before BM25 anchors and hybrid fill; \emph{planner-rerank} packs planner-selected atoms before cross-encoder reranked fill; \emph{Planner+BM25} uses BM25 fill. We also report an earlier-adapter sensitivity check.

\paragraph{Routers, statistics, and artifacts.}
An oracle router chooses post hoc using gold correctness and measures complementarity rather than deployable performance. Leakage-safe routers choose among context-construction policies using only inference-available features under stratified 5-fold and leave-domain-out validation. We report accuracy, paired wins/losses, exact two-sided sign-test $p$-values, and 95\% intervals. The public repository provides evaluation and analysis code, portable configurations, split IDs, environment records, checksums, text-free per-example predictions, atom IDs, retrieval/reranker ranks and scores, planner-selected and packed IDs, and paired-test outputs.\footnote{\url{https://github.com/aapplepku/context-planning-vs-retrieval}} Released scripts regenerate the main, split, domain, packing-order, paired-sign, and compact mechanism analyses; router and selector results and some intervals are provided as audited summaries. Raw LongBench-v2 contexts, base-model weights, and adapter weights are not redistributed. We do not claim bitwise reproduction of historical GPU inference, and latency is descriptive only.

\section{Results}
\subsection{Direct Planning Loses to Retrieval}
Table~\ref{tab:full} gives the full-budget result. Anchored hybrid retrieval is best at 36.18\%, and BM25 is nearly tied at 35.98\%. The best direct planner-guided method reaches 34.19\%. Planner+BM25 is lower than BM25 by 3.78 points (56 BM25-only wins versus 37 planner-only wins; $p=0.061$). Thus, direct planner-guided methods do not beat strong retrieval in this setup.

\begin{table}[t]
\centering\small
\begin{tabular}{lrrr}
\toprule
Method & Acc. & 95\% CI & Pair $p$ \\
\midrule
Hybrid anchor & 36.18 & [32.10, 40.47] & -- \\
BM25 & 35.98 & [31.91, 40.27] & 1.000$^{a}$ \\
Planner+hybrid anchor & 34.19 & [30.18, 38.45] & 0.399$^{b}$ \\
Planner+BM25 & 32.21 & [28.27, 36.41] & 0.061$^{c}$ \\
Dense & 30.22 & [26.37, 34.37] & -- \\
\bottomrule
\end{tabular}
\caption{Full 18k-context results on 503 questions. CIs are Wilson intervals. $^{a}$BM25 vs. hybrid anchor. $^{b}$Planner+hybrid anchor vs. hybrid anchor. $^{c}$Planner+BM25 vs. BM25.}
\label{tab:full}
\end{table}

\paragraph{Held-out robustness.}
On the 152 questions with no SFT train/development overlap, anchored hybrid reaches 42.11\% and planner-guided anchored hybrid reaches 36.84\% (difference $-5.26$ points; 12 planner wins, 20 losses, 120 ties; $p=0.215$; paired-bootstrap 95\% CI $[-12.50,1.97]$ points). This lower-powered subset does not provide evidence that planning robustly exceeds the retrieval control.

\begin{table}[t]
\centering\scriptsize
\setlength{\tabcolsep}{2.5pt}
\begin{tabular}{lrrrr}
\toprule
Domain ($n$) & BM25 & Anchor & P+anchor & $\Delta$ \\
\midrule
Code repo. (50) & 40.00 & 40.00 & 34.00 & -6.00 \\
Long ICL (81) & 34.57 & 34.57 & 35.80 & +1.23 \\
Structured (33) & 39.39 & 39.39 & 36.36 & -3.03 \\
Dialogue hist. (39) & 25.64 & 25.64 & 25.64 & 0.00 \\
Multi-doc QA (125) & 30.40 & 31.20 & 30.40 & -0.80 \\
Single-doc QA (175) & 41.14 & 41.14 & 37.71 & -3.43 \\
\bottomrule
\end{tabular}
\caption{Full-budget accuracy (\%) by LongBench-v2 domain. $\Delta$ is planner-guided anchored hybrid minus anchored hybrid. These subgroup results are descriptive.}
\label{tab:domain}
\end{table}

At full budget, planner-guided anchored hybrid exceeds its non-planner counterpart in one of six domains, ties in one, and trails in four (Table~\ref{tab:domain}). Several domain subsets are small, so we do not treat these differences as confirmatory.

\subsection{Oracle Routing Does Not Become a Router}
Oracle routing shows complementary errors: an oracle over hybrid anchor and planner+hybrid reaches 46.52\%, and an all-method oracle reaches 56.26\%. These are post-hoc upper bounds, not deployable policies. Table~\ref{tab:router} separates them from leakage-safe routers using task/domain metadata, context-length statistics, retrieval-score summaries, planner parse/selection metadata, and choice-confidence features. Across stratified 5-fold and leave-domain-out validation, logistic regression, shallow random forests, histogram gradient boosting, domain-only rules, and default-train baselines fail to improve over the best single method.

\begin{table}[t]
\centering\small
\begin{tabularx}{0.94\linewidth}{Yrrr}
\toprule
Policy & Acc. & Gain & Prot. \\
\midrule
Best single: hybrid anchor & 36.18 & 0.00 & fixed \\
Oracle: hybrid/planner & 46.52 & +10.34 & post hoc \\
Oracle: all methods & 56.26 & +20.08 & post hoc \\
\midrule
Pair routers & 36.18 & 0.00 & 5-fold/LDO \\
BM25-vs-planner routers & 35.98 & -0.20 & 5-fold/LDO \\
Five-way routers & 35.98 & -0.20 & 5-fold/LDO \\
Domain router & 36.38 & +0.20 & same-data \\
\bottomrule
\end{tabularx}
\caption{Oracle bounds and routers. Fixed = best single method; post hoc = oracle using gold correctness; 5-fold/LDO = stratified 5-fold and leave-domain-out validation; same-data = optimistic in-sample rule. The domain router selects, per domain, the better of hybrid anchor and planner-guided anchored hybrid using in-sample accuracy; its +0.20-point gain corresponds to one additional question.}
\label{tab:router}
\end{table}

\subsection{Budget and Rerank Controls Erase Most Gains}
A weaker 6k comparison initially favored plan-guided hybrid over anchored hybrid by 1.99 points. We therefore add stronger budgeted selectors and reranker controls. Table~\ref{tab:budgetrerank} reports the decisive comparisons. Against the best same-budget non-learned selector, the 6k planner gain is only +0.40 points with a wide interval and $p=0.930$; at 9k the planner loses by 1.19 points. Against strong non-planner rerank controls, planner-rerank has a +1.79-point estimate at 6k, but the paired interval crosses zero and $p=0.478$. At 9k, it ties the best non-planner reranker.

\begin{table}[t]
\centering\small
\setlength{\tabcolsep}{3.2pt}
\begin{tabular}{llrrl}
\toprule
Set & Bgt & Ctrl & Plan & $\Delta$ [95\% CI] \\
\midrule
Sel. & 6k & 32.80 & 33.20 & +0.40 [-4.01,+4.81] \\
Sel. & 9k & 35.59 & 34.39 & -1.19 [-5.43,+3.04] \\
Rerank & 6k & 31.41 & 33.20 & +1.79 [-2.58,+6.16] \\
Rerank & 9k & 34.19 & 34.19 & 0.00 [-4.37,+4.17] \\
\bottomrule
\end{tabular}
\caption{Best planner method versus best non-planner control at each budget. Paired $p$-values are 0.930, 0.645, 0.478, and 1.000 in row order. Values are percentage points.}
\label{tab:budgetrerank}
\end{table}

The earlier sensitivity adapter raises the 6k planner-rerank score to 34.79\%, but still lacks paired support ($p=0.146$; CI $[-0.99,+7.75]$ points) and does not beat the 9k control. Its selected-atom Jaccard with the final adapter is 0.881, indicating similar selections but not a robust accuracy gain.

\subsection{Selection, Order, and Score Mechanisms}
Holding the planner-selected set fixed, hybrid-retrieval order changes accuracy relative to planner order by $-1.19$ points at 6k ($p=0.461$) and $+0.99$ points at 9k ($p=0.511$). The reversal does not support a systematic packing-order advantage. We also compare planner-selected IDs with BM25 and cross-encoder rankings, measure high-ranked-atom displacement, and test four predeclared score-flatness statistics. Planner--reranker Jaccard@10 is 0.084 for planner wins versus 0.097 for losses at 6k, and 0.082 versus 0.111 at 9k; planner contexts exclude an average 8.5--9.3 of the reranker's top-10 atoms. None of 92 win--loss tests remains significant after Benjamini--Hochberg correction. Thus, the outputs do not support a stable claim that planner wins occur specifically when retrieval scores are flatter. These are rank and context-membership analyses, not evidence-recall measurements, because LongBench-v2 provides no gold atom-level evidence labels for this study.

\section{Discussion and Conclusion}
When compared only with prompted or heuristic planning and direct context baselines, a learned planner can look useful. After adding BM25, anchored hybrid packing, and reranking controls, most apparent gains disappear, even though the planner remains trainable and sometimes changes the evidence set helpfully. Paired analysis is essential: the +1.79-point 6k estimate comes from 68 planner wins and 59 losses among 503 questions. Under the observed discordance, confirming a +1.5-point paired effect would require roughly 8,800 examples and a +2.0-point effect roughly 4,900 examples.

In this bounded LongBench-v2 MCQ study, learned context planning is better viewed as a weak relevance feature than as a replacement for retrieval. Future studies should compare against anchored hybrid retrieval, budgeted selectors, leakage-safe routing, and strong rerankers before interpreting gains as evidence of planning ability.

\section*{Limitations}
Our conclusions are restricted to LongBench-v2 MCQ, Qwen2.5-7B-Instruct, one outcome-supervised planner recipe, one overlapping character-window atomization scheme, the evaluated lexical/dense retrievers, and the MiniLM cross-encoder. The all-503 diagnostic is partly transductive for the 168 questions represented in SFT train/development data; the disjoint test analysis has only 152 questions and limited power for small paired effects. We do not evaluate open-ended QA, other model scales, alternative supervision sources, or real-world deployment, and we have no external replication. A larger or differently mined trace set could change the outcome. The historical run predates strict locked-run enforcement and immutable model-revision pinning, and latency is descriptive rather than a portable efficiency claim. LongBench-v2 supplies no gold atom-level evidence labels for this analysis, so mechanism results use atom IDs, ranks, scores, and displacement rather than evidence recall. These results do not imply that planners are generally ineffective; they show that this planner did not robustly beat strong retrieval controls under the studied conditions.

\section*{Acknowledgments}
We thank the Insights 2026 reviewers for their careful comments and suggestions. Generative AI assistants were used to help draft and revise portions of this manuscript and to help prepare the analysis and artifact-release scripts. All experimental results, statistics, and claims were verified by the authors against the released artifacts, and the authors take full responsibility for the content.

\end{document}